\documentclass[conference,review]{IEEEtran}
\IEEEoverridecommandlockouts

\usepackage{cite}
\usepackage{amsmath,amssymb,amsfonts}
\usepackage{algorithmic}
\usepackage{graphicx}
\usepackage{textcomp}
\usepackage{soul}
\usepackage{cuted}
\usepackage{xcolor}
\usepackage{float}
\usepackage{dblfloatfix} 
\usepackage{hyperref}
\usepackage{caption}

\usepackage{placeins}

\def\BibTeX{{\rm B\kern-.05em{\sc i\kern-.025em b}\kern-.08em
    T\kern-.1667em\lower.7ex\hbox{E}\kern-.125emX}}
\begin{document}

\title{MS2Mesh-XR: Multi-modal Sketch-to-Mesh Generation in XR Environments

\thanks{This work was supported by the Research Grants Council of the Hong Kong Special Administrative Region, China (Project No.: T45-401/22-N); and in part by The Chinese University of Hong Kong (Project No.: 4055212).}
}

\author{\IEEEauthorblockN{{Yuqi Tong$^{*}$, Yue Qiu$^{*}$, Ruiyang Li, Shi Qiu$^{\dagger}$, Pheng-Ann Heng}}\\
\IEEEauthorblockA{\textit{Department of Computer Science and Engineering, The Chinese University of Hong Kong, HKSAR China} \\
\textit{Institute of Medical Intelligence and XR, The Chinese University of Hong Kong, HKSAR China}
}
\thanks{$^{*}$ denotes equal contribution.}
\thanks{$^{\dagger}$ corresponding author: shiqiu@cse.cuhk.edu.hk}
}

\maketitle

\begin{strip}
\begin{minipage}{\textwidth}\centering
\vspace{-60pt}
\includegraphics[width=0.85\textwidth]{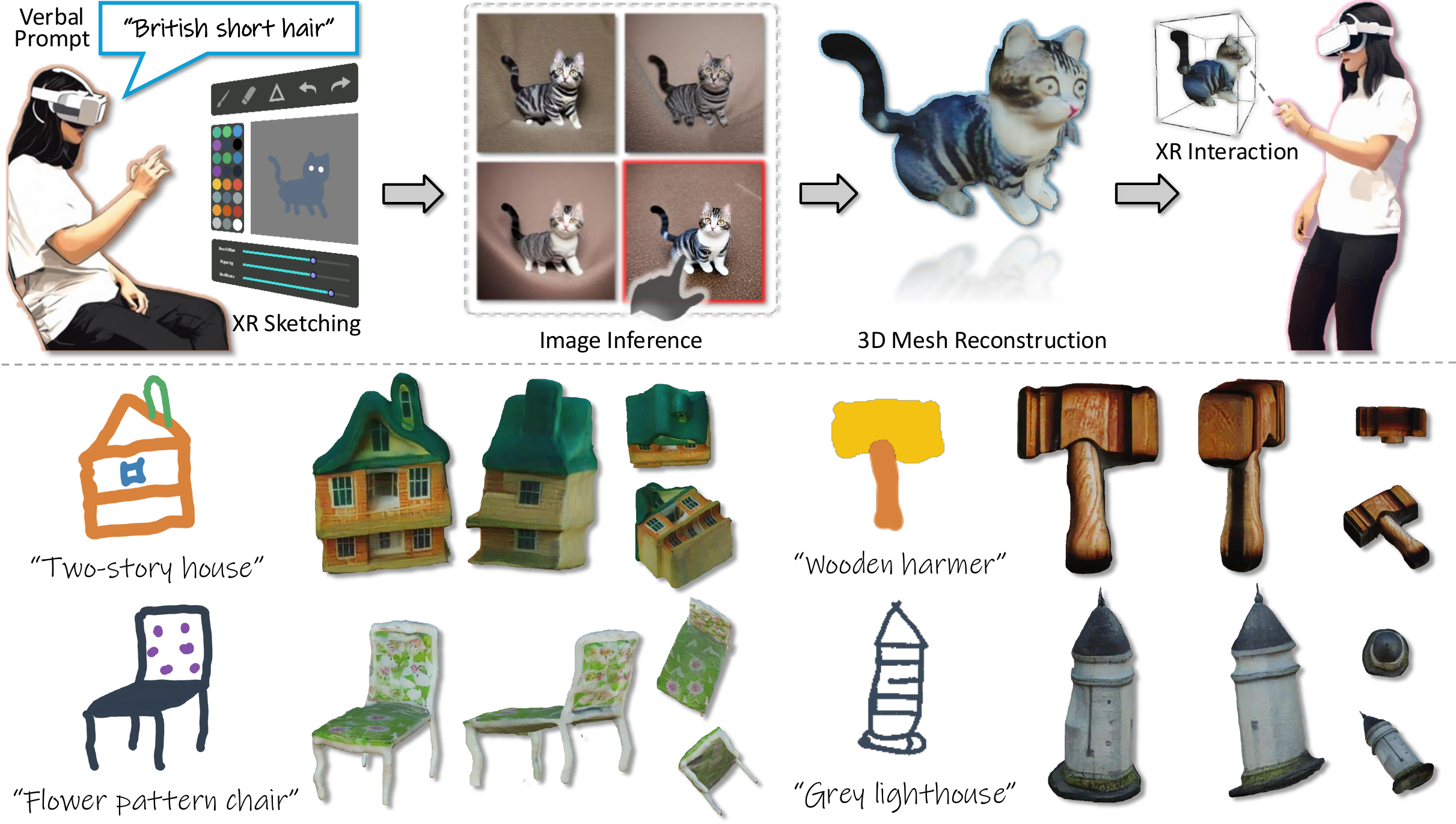}
\captionof{figure}{MS2Mesh-XR integrates hand-drawn sketches with voice inputs to rapidly generate realistic 3D meshes for natural user interactions in XR environments.}
\vspace{-10pt}
\label{teaser}
\end{minipage}
\end{strip}

\begin{abstract}
We present \emph{MS2Mesh-XR}, a novel multi-modal sketch-to-mesh generation pipeline that enables users to create realistic 3D objects in extended reality (XR) environments using hand-drawn sketches assisted by voice inputs. In specific, users can intuitively sketch objects using natural hand movements in mid-air within a virtual environment. By integrating voice inputs, we devise ControlNet to infer realistic images based on the drawn sketches and interpreted text prompts. Users can then review and select their preferred image, which is subsequently reconstructed into a detailed 3D mesh using the Convolutional Reconstruction Model. In particular, our proposed pipeline can generate a high-quality 3D mesh in less than 20 seconds, allowing for immersive visualization and manipulation in run-time XR scenes. We demonstrate the practicability of our pipeline through two use cases in XR settings. By leveraging natural user inputs and cutting-edge generative AI capabilities, our approach can significantly facilitate XR-based creative production and enhance user experiences. Our code and demo will be available at: \url{https://yueqiu0911.github.io/MS2Mesh-XR/}.

\end{abstract}


\section{Introduction}

Automated 3D content generation in extended reality (XR) environments has gained significant attention. However, creating high-quality 3D objects that support dynamic, immersive visualization and interaction remains a major challenge, requiring extensive time, effort, and expertise~\cite{zhu2024ssp, cui2022energy}. Existing approaches~\cite{jackson2016liftoff,michel2021text2mesh,anonymous2021stripbrush,luo2022structure} mainly rely on XR-based sketching to create 3D objects, while they are also confronted with two practical issues: (i) they often require users to possess advanced drawing skills; (ii) the inherent inaccuracy of interactive painting in XR scenes prevents users from creating high-fidelity 3D models, especially those with fine details.

Recent advancements in AI-generated content (AIGC) have enabled the creation of 3D content from text~\cite{liang2024luciddreamer,li2024dreamscene,zhou2025dreamscene360} or images \cite{hong2024lrm, wang2024crm, wei2024meshlrm}. However, their practical application in achieving interactive and personalized XR experiences remains largely unexplored. To address this gap, we introduce an innovative pipeline, \textbf{MS2Mesh-XR} (\textbf{M}ulti-modal \textbf{S}ketch-\textbf{to-Mesh} in \textbf{XR}), to assist users in intuitively creating high-fidelity 3D objects in XR environments using natural interaction inputs. 

Our pipeline in Fig.~\ref{teaser} fully supports editable and customizable 3D content generation in XR by exploiting multiple modalities of information provided by users, including hand-drawn sketches and voice inputs. Specifically, we first capture multi-modal information from the user, integrating the geometric context extracted from hand-drawn sketches along with the text prompts interpreted from voice inputs. Then, this multi-modal information is leveraged to infer a high-fidelity image via the ControlNet\cite{zhang2023adding}, which serves as the basis for further processing. Next, the inferred image is processed by the Convolution Reconstruction Model\cite{wang2024crm}, where a multi-view image diffusion model produces six orthographic images for the final reconstruction of a textured mesh. 
Our proposed multi-modal sketch-to-mesh generation pipeline allows for user-centered creation and manipulation of high-quality textured 3D mesh models based on natural user inputs. The entire generation process takes no more than 20 seconds, producing a high-quality mesh that can be immediately imported and interacted with in run-time XR scenes. Moreover, we demonstrate use cases for our proposed pipeline in both virtual reality (VR) and mixed reality (MR) modes. Our approach not only generates photorealistic 3D models that enhance immersive experiences but also supports natural user interactions, boosting user engagement and creativity in XR-based 3D object creation scenarios. 

Our key contributions are summarized as follows:
\begin{itemize}
    \item We develop a mesh generation pipeline that leverages ControlNet and Convolutional Reconstruction Model to create high-quality 3D mesh models.
    \item We integrate multi-modal natural user inputs, including hand-drawn sketches and voice prompts, to accurately capture user intentions via intuitive XR-based interactions.
    \item We demonstrate two use cases, asset creation in VR mode and interior design in MR mode, showcasing the effectiveness and applicability of our pipeline across different XR scenarios.

\end{itemize}

\section{Related Works}

\begin{figure*}[t]
\centerline{ \includegraphics[width=\textwidth]{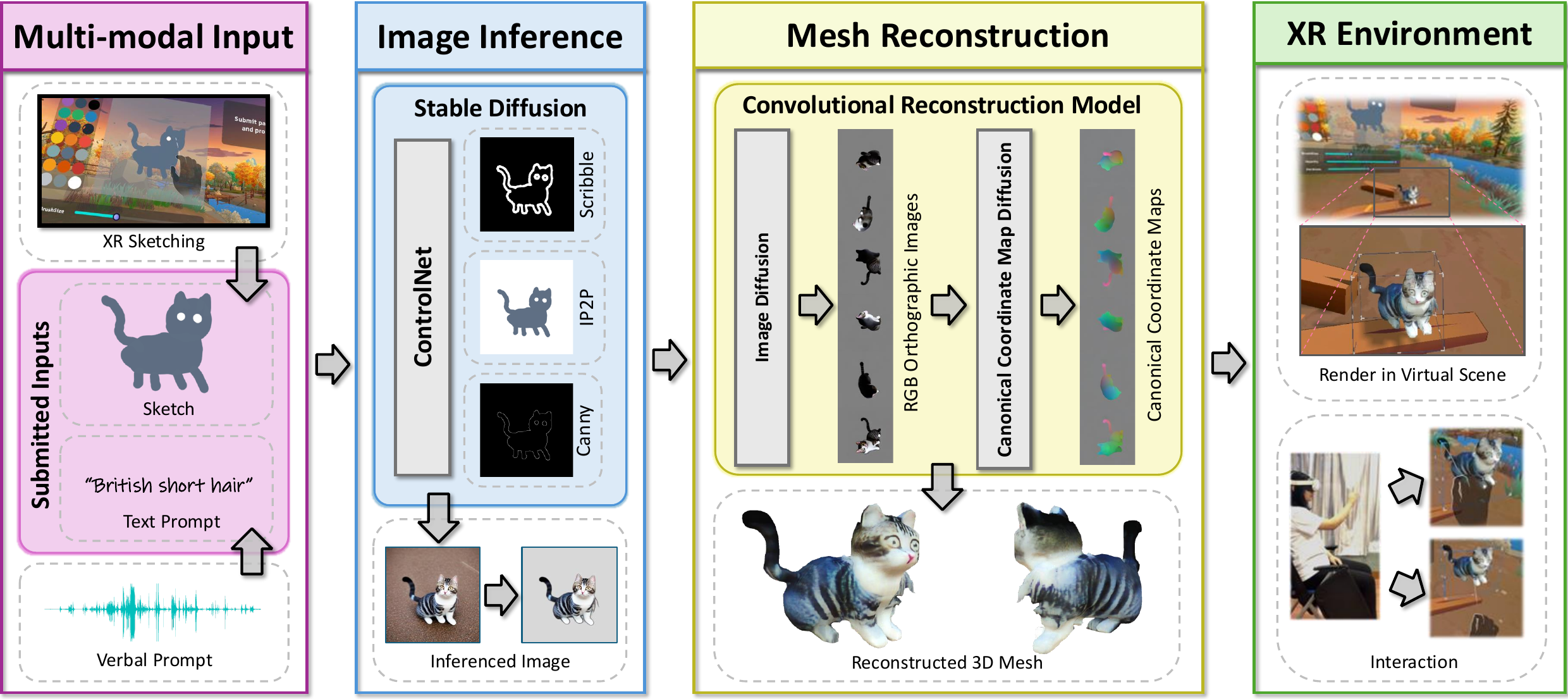}}
\caption{Overview for our MS2Mesh-XR pipeline. Multi-modal inputs from user sketch and voice feed into the image inference module, which generates a reference image. The mesh reconstruction module then uses the reference image to reconstruct a corresponding 3D mesh, leveraging multiview images generated by the diffusion models. The generated 3D object is finally rendered in the XR environment supported by intuitive user interactions.}
\vspace{-10pt}
\label{pipeline}
\end{figure*}

\noindent\textbf{Text/Sketch-to-Mesh Generation.} Dreamfusion \cite{poole2022dreamfusion} proposes Score Distillation Sampling (SDS), a novel technique that distills 3D assets from pre-trained 2D text-to-image diffusion models. SDS enables effective 3D model training by identifying specific modes in the text-guided diffusion process, allowing 2D diffusion model knowledge to be transferred into 3D generation. This method inspires a substantial body of follow-up research \cite{chen2023fantasia3d, huang2023dreamwaltz, lin2023magic3d, metzer2023latentnerf, poole2022dreamfusion, qian2023magic123, zhang2023avatarverse, liang2024luciddreamer}, and has become a pivotal component in the text-to-3D generation domain.
More recently, works in 3D generation focus on improving the editability of models. Sketch2Scene~\cite{Sketch2Scene2024} enables the automatic generation of interactive 3D game scenes from simple user sketches. WorldSmith~\cite{WorldSmith2023} provides a tool for users to iteratively build and modify fictional worlds through prompts and generative AI, making the world-building process more flexible. Magic3DSketch~\cite{Magic3DSketch2024} enhances sketch-based 3D modeling by using language-image pre-training, enabling the creation of highly detailed and customizable 3D models from sparse sketches, with the ability to refine features based on text input.

\noindent\textbf{AIGC in XR.} 
Advances in AI-generated content (AIGC) have significantly impacted XR applications, particularly in 3D content creation and user interaction~\cite{zhu2025pcf, qiu2024advancing}. GAN-based 3D VR sketch synthesis\cite{bdiot2023synthesizing} automates the generation of 3D sketches, simplifying the 3D modeling process and making it accessible to non-experts. VRCopilot~\cite{arxiv2408vrcopilot} extends this by combining generative AI models with user input to co-create 3D layouts or generate VR scenes directly from text descriptions, reducing manual effort and enhancing control in immersive environments~\cite{vr2024text2vrscene}. 
Some approaches enhance AR interactions with voice-driven control of AI-generated images and text, offering a sketch-based interface for creating and refining content, thereby making the design process more natural and user-friendly\cite{arxiv2023speechtoimage,arxiv2020arsketch}. 
Additionally, other works focus on real-time human-AI collaboration for tasks like fashion design and 3D modeling, emphasizing workflow efficiency and intuitive interaction through sketching and reference imagery~\cite{vr2018fashiondesign, tvcg2016liftoff}. Different from existing works, we perform a systematic integration of voice commands and editable hand-drawing sketches for AIGC-assisted 3D mesh generation. By leveraging these multi-modal user inputs, our proposed method achieves greater flexibility and efficiency in 3D content creation, relying solely on natural and intuitive XR-based interactions.

\section{Methodology}

Fig. \ref{pipeline} outlines our proposed pipeline for 3D content creation and manipulation in XR environments. The workflow begins with users' free-hand sketching to define the basic shape and structure of the desired 3D object. Then, users can further refine the design by specifying additional details through verbal prompts. These multi-modal inputs are processed by the image inference module powered by the ControlNet\cite{zhang2023adding}, which infers a high-fidelity image aligning with the user’s expectations. Another key component of our proposed pipeline is the mesh reconstruction module powered by Convolutional Reconstruction Model \cite{wang2024crm}, which uses the generated image as a reference to reconstruct a detailed 3D mesh model accordingly. This mesh reconstruction process involves several stages, including image diffusion, 3D reconstruction, and mesh refinement. Finally, the generated 3D model can be imported into a run-time XR environment, allowing users to intuitively observe and interact with the model in an immersive context.

\subsection{Multi-modal Input in XR: Sketch and Voice}

Our method combines user freehand sketch and voice inputs to extract multi-modal information that captures the main features of the desired 3D mesh output. Practically, we record user sketches by employing a dedicated XR-based virtual canvas, coupled with the Unity platform's asset 2D/3D Paint\cite{unity2d3dpaint}.
To facilitate freehand painting in XR mode, we use the MRTK3~\cite{mrtk} hand ray, which interacts with the virtual drawing board by registering painting actions at the hit point.
In addition, user voice data is transcribed into text prompts using the Meta Voice SDK\cite{meta_voice_sdk}. By integrating both visual and verbal information, the captured multi-modal data effectively represents user intentions.

\begin{figure*}[!t]
\centerline{ \includegraphics[width=0.99\textwidth]{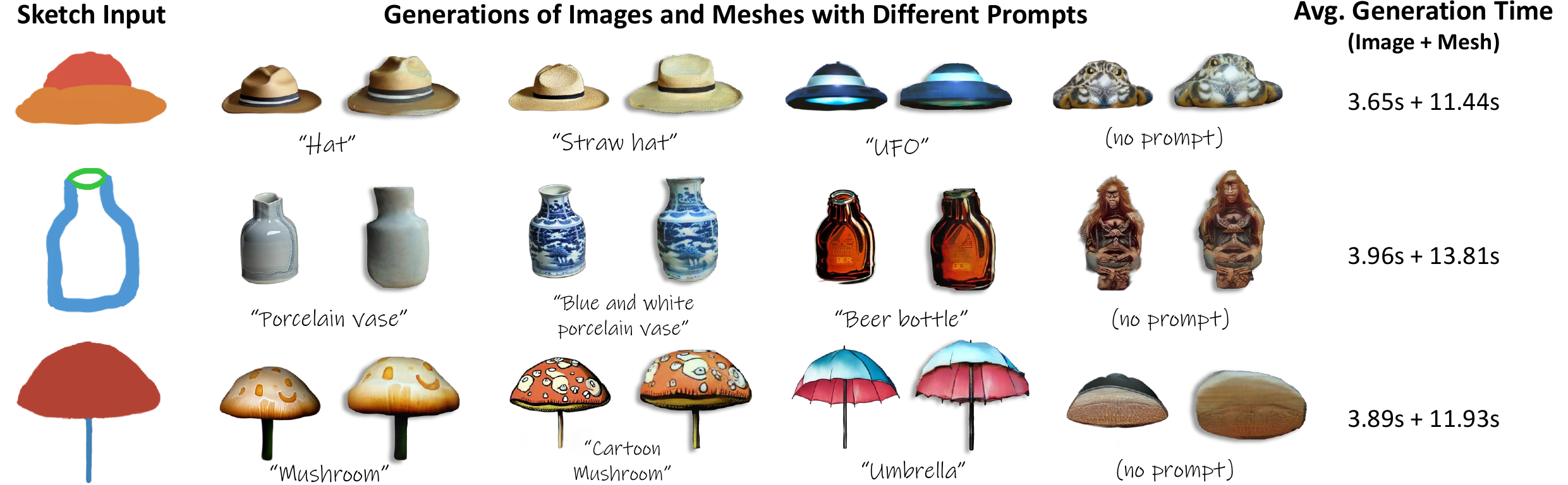}}
\caption{Comparison of images and meshes generated by sketches with different prompts. For each pair of generation results, the left one shows the generated 2D image, while the right one presents the corresponding generated 3D mesh. }
\label{result}
\end{figure*}

\subsection{Image Inference}

In our work, we adopt ControlNet 1.1 \cite{zhang2023adding} as a robust framework to translate multi-modal user inputs into a meaningful and high-fidelity output image. 
Theoretically, ControlNet utilizes a pre-trained text-to-image diffusion model, incorporating additional spatial conditioning during the denoising process. In practice, we utilize three distinct models of ControlNet, including Scribble, Canny, and IP2P, to enhance the quality of the inferred 2D image content. Particularly, the IP2P and Scribble models handle the generation of primary shapes based on sketch outlines, while the Canny model refines the image by adding detailed features.

By using both freehand sketches drawn by the user and text prompts derived from voice inputs, our image inference module generates a high-fidelity image that closely matches the user’s intentions. The inferred image is also processed through the Python-based rembg tool \cite{gatis2023rembg} for background removal, yielding a clear, object-centered output for subsequent image-to-mesh reconstruction operations.

\subsection{Mesh Reconstruction}
Following the image inference process, the mesh reconstruction module utilizes the advanced Convolutional Reconstruction Model (CRM) \cite{wang2024crm} to generate a corresponding high-quality 3D mesh. This process begins with multi-view diffusion models, which produce six orthographic images and canonical coordinate maps (CCMs). These six orthographic views, along with their corresponding CCMs, are then merged to form expanded images for spatial alignment across all input data. Next, the expanded images are processed by a convolutional U-Net, which maps the images and CCMs to triplane-based representations. These triplane features are subsequently decoded by three compact Multi-Layer Perceptrons (MLPs) to extract signed distance function (SDF) values, color, and Flexicube parameters. Finally, the dual marching cubes algorithm \cite{schaefer2005dual} processes the SDF values, color, and Flexicube parameters to reconstruct a textured 3D mesh. In practice, the structured workflow described in \cite{wang2024crm} reconstructs high-resolution, detailed 3D meshes for given images, particularly well-suited for XR applications.

\subsection{Integration into XR}

To integrate the generated 3D mesh into an XR scene running on a head-mounted device (HMD), we transfer the mesh object, along with its associated material and texture maps, using the Hypertext Transfer Protocol (HTTP) over a local network to ensure stable transmission. A runtime OBJ importer for Unity3D is then used to incorporate the textured mesh into the scene. Moreover, to facilitate intuitive interaction, we apply the ``Object Manipulator'' and ``Bounds Control'' components from the MRTK3 toolkit \cite{mrtk}, allowing users to manipulate the object within the XR environment through a bounding box interface.

\section{Results}

\subsection{Implementation Details }
The MS2Mesh-XR pipeline comprises two main components. For the image inference module, we use three distinct ControlNet models with specific weights: Scribble (0.55), Canny (0.05), and IP2P (0.5). For the mesh reconstruction module, we use the fine-tuned Convolutional Reconstruction Model\cite{wang2024crm} with approximately 300M parameters, including a U-Net with channels [64, 128, 128, 256, 256, 512, 512],  attention blocks at resolutions [32, 16, 8], and the Flexicubes grid size of 80 are utilized. We develop our pipeline using Unity3D and Meta Quest 3. For data processing, we use a workstation with an RTX 4090 GPU for image inference and mesh generation operations, while data transfers between the HMD and workstation are managed via HTTP.

\subsection{Qualitative Evaluation }

Our method supports the generation of high-quality 3D mesh models from sketches and voice inputs in XR environments. We also provide sufficient flexibility in our generation pipeline based on different voice inputs. As demonstrated in Fig. \ref{result}, the average generation time is approximately 3.83 seconds for images and 12.39 seconds for meshes. Given the same hand-drawn sketches, users can utilize different verbal prompts to generate diverse and realistic results. Sketches and texts clearly provide complementary information: sketches define the general shape and geometry, while texts specify detailed and representative features. Without semantic guidance from text-based prompts, the generated mesh may exhibit less meaningful global shape and local structures.

\section{Use Cases}

We apply our pipeline for two distinct use cases: the first involves an immersive VR scene where users can create interactive assets, while the second utilizes an MR scenario that allows users to decorate a real-world space with self-generated 3D furniture. These examples demonstrate how our method effectively supports 3D content creation. More details about these two use cases are presented in our demo video.


\subsection{Interactive Assets Creation in VR}
In VR,  our MS2Mesh-XR pipeline transforms the 2D sketch and associated voice prompts into a dedicated 3D asset in real time.
Users can then apply backend-managed functions to trigger different effects for the specified items (such as ``sparkle'' or ``smoke''), This approach shows potential use in VR games, such as scene arrangement, avatar asset generation, and creative sandbox games, offering a novel way to enhance interactivity and customization in virtual environments~\cite{anonymous2023synthesizing}.
\subsection{Interior Design in MR}

In an MR environment, MS2Mesh-XR empowers users to create personalized indoor designs by drawing furniture outlines and refining features (\emph{e.g.}, color, material, pattern) using voice. Users begin by leveraging a multi-modal interface to create furniture models (\emph{e.g.}, tables, chairs, sofas), specifying dimensions and aesthetic details. Freely arranging these generated models around a real-world room, users can adjust furniture sizes, reposition items, and refine design elements based on immediate visual feedback. Once satisfied with the layout, users can save and export their designs, obtaining a tailored 3D model with a unique style and optimized layout. 


\section{Conclusion}

We propose an innovative 3D content creation pipeline, MS2Mesh-XR, which integrates hand-drawn sketches with voice inputs, leveraging generative AI tools such as ControlNet for high-fidelity image inference and the Convolutional Reconstruction Model for realistic 3D mesh generation. This novel pipeline supports the creation of highly detailed textured meshes in XR environments and offers an adaptive solution for real-time 3D creation in VR/MR applications, enhancing user experience through intuitive interactions and real-time visualizations. However, our approach has limitations: for example, the line colors in the sketch do not map well to the 3D model due to ControlNet constraints, and the pipeline is restricted by the capabilities of deployed algorithms and GPU device. In the future, we aim to overcome these challenges by integrating more advanced and efficient methods to enhance both performance and accuracy.


\bibliographystyle{IEEEtran}
\bibliography{ref}

\begin{thebibliography}{10}
\providecommand{\url}[1]{#1}
\csname url@samestyle\endcsname
\providecommand{\newblock}{\relax}
\providecommand{\bibinfo}[2]{#2}
\providecommand{\BIBentrySTDinterwordspacing}{\spaceskip=0pt\relax}
\providecommand{\BIBentryALTinterwordstretchfactor}{4}
\providecommand{\BIBentryALTinterwordspacing}{\spaceskip=\fontdimen2\font plus
\BIBentryALTinterwordstretchfactor\fontdimen3\font minus \fontdimen4\font\relax}
\providecommand{\BIBforeignlanguage}[2]{{%
\expandafter\ifx\csname l@#1\endcsname\relax
\typeout{** WARNING: IEEEtran.bst: No hyphenation pattern has been}%
\typeout{** loaded for the language `#1'. Using the pattern for}%
\typeout{** the default language instead.}%
\else
\language=\csname l@#1\endcsname
\fi
#2}}
\providecommand{\BIBdecl}{\relax}
\BIBdecl

\bibitem{zhu2024ssp}
R.~Zhu, D.~Kang, K.-H. Hui, Y.~Qian, S.~Qiu, Z.~Dong, L.~Bao, P.-A. Heng, and C.-W. Fu, ``Ssp: Semi-signed prioritized neural fitting for surface reconstruction from unoriented point clouds,'' in \emph{WACV}, 2024, pp. 3769--3778.

\bibitem{cui2022energy}
R.~Cui, S.~Qiu, S.~Anwar, J.~Zhang, and N.~Barnes, ``Energy-based residual latent transport for unsupervised point cloud completion,'' \emph{BMVC}, 2022.

\bibitem{jackson2016liftoff}
B.~Jackson and D.~F. Keefe, ``Lift-off: Using reference imagery and freehand sketching to create 3d models in vr,'' in \emph{2016 IEEE Transactions on Visualization and Computer Graphics}, vol.~22, no.~4.\hskip 1em plus 0.5em minus 0.4em\relax IEEE, 2016, pp. 1442--1451.

\bibitem{michel2021text2mesh}
O.~Michel, R.~Bar-On, R.~Liu, S.~Benaim, and R.~Hanocka, ``Text2mesh: Text-driven neural stylization for meshes,'' \emph{arXiv preprint arXiv:2112.03221}, 2021.

\bibitem{anonymous2021stripbrush}
Anonymous, ``Stripbrush: A constraint-relaxed 3d brush reduces physical effort and enhances the quality of spatial drawing,'' \emph{arXiv preprint arXiv:2109.03845}, 2021.

\bibitem{luo2022structure}
L.~Luo, Y.~Gryaditskaya, Y.~Yang, T.~Xiang, and Y.-Z. Song, ``Structure-aware 3d vr sketch to 3d shape retrieval,'' \emph{arXiv preprint arXiv:2209.10008}, 2022.

\bibitem{liang2024luciddreamer}
Y.~Liang, X.~Yang, J.~Lin, H.~Li, X.~Xu, and Y.~Chen, ``Luciddreamer: Towards high-fidelity text-to-3d generation via interval score matching,'' in \emph{Proceedings of the IEEE/CVF Conference on Computer Vision and Pattern Recognition (CVPR)}.\hskip 1em plus 0.5em minus 0.4em\relax The Computer Vision Foundation, 2024.

\bibitem{li2024dreamscene}
H.~Li, H.~Shi, W.~Zhang, W.~Wu, Y.~Liao, L.~Wang, L.-h. Lee, and P.~Zhou, ``Dreamscene: 3d gaussian-based text-to-3d scene generation via formation pattern sampling,'' \emph{arXiv preprint arXiv:2404.03575}, 2024.

\bibitem{zhou2025dreamscene360}
S.~Zhou, Z.~Fan, D.~Xu, H.~Chang, P.~Chari, T.~Bharadwaj, S.~You, Z.~Wang, and A.~Kadambi, ``Dreamscene360: Unconstrained text-to-3d scene generation with panoramic gaussian splatting,'' in \emph{European Conference on Computer Vision}.\hskip 1em plus 0.5em minus 0.4em\relax Springer, 2025, pp. 324--342.

\bibitem{hong2024lrm}
Y.~Hong, K.~Zhang, J.~Gu, S.~Bi, Y.~Zhou, D.~Liu, F.~Liu, T.~Sunkavalli, K.~Bui, and H.~Tan, ``Lrm: Large reconstruction model for single image to 3d,'' in \emph{ICLR}, 2024.

\bibitem{wang2024crm}
Z.~Wang, Y.~Wang, Y.~Chen, C.~Xiang, S.~Chen, D.~Yu, C.~Li, H.~Su, and J.~Zhu, ``Crm: Single image to 3d textured mesh with convolutional reconstruction model,'' \emph{arXiv preprint arXiv:2403.05034}, 2024.

\bibitem{wei2024meshlrm}
X.~Wei, K.~Zhang, S.~Bi, H.~Tan, F.~Luan, V.~Deschaintre, K.~Sunkavalli, H.~Su, and Z.~Xu, ``Meshlrm: Large reconstruction model for high-quality mesh,'' \emph{arXiv preprint arXiv:2404.12385}, 2024.

\bibitem{zhang2023adding}
L.~Zhang, A.~Rao, and M.~Agrawala, ``Adding conditional control to text-to-image diffusion models,'' \emph{arXiv preprint arXiv:2302.05543}, 2023.

\bibitem{poole2022dreamfusion}
B.~Poole, A.~Jain, J.~T. Barron, and B.~Mildenhall, ``Dreamfusion: Text-to-3d using 2d diffusion,'' \emph{arXiv preprint arXiv:2209.14988}, 2022.

\bibitem{chen2023fantasia3d}
R.~Chen, Y.~Chen, N.~Jiao, and K.~Jia, ``Fantasia3d: Disentangling geometry and appearance for high-quality text-to-3d content creation,'' in \emph{International Conference on Computer Vision (ICCV)}, 2023.

\bibitem{huang2023dreamwaltz}
Y.~Huang, J.~Wang, A.~Zeng, H.~Cao, X.~Qi, Y.~Shi, Z.-J. Zha, and L.~Zhang, ``Dreamwaltz: Make a scene with complex 3d animatable avatars,'' \emph{arXiv preprint arXiv:2305.02463}, 2023.

\bibitem{lin2023magic3d}
C.-H. Lin, J.~Gao, L.~Tang, T.~Takikawa, X.~Zeng, X.~Huang, K.~Kreis, S.~Fidler, M.-Y. Liu, and T.-Y. Lin, ``Magic3d: High-resolution text-to-3d content creation,'' in \emph{IEEE/CVF Conference on Computer Vision and Pattern Recognition (CVPR)}, 2023.

\bibitem{metzer2023latentnerf}
G.~Metzer, E.~Richardson, O.~Patashnik, R.~Giryes, and D.~Cohen-Or, ``Latent-nerf for shape-guided generation of 3d shapes and textures,'' in \emph{IEEE/CVF Conference on Computer Vision and Pattern Recognition (CVPR)}, 2023.

\bibitem{qian2023magic123}
G.~Qian, J.~Mai, A.~Hamdi, J.~Ren, A.~Siarohin, B.~Li, H.-Y. Lee, P.~Wonka, S.~Tulyakov \emph{et~al.}, ``Magic123: One image to high-quality 3d object generation using both 2d and 3d diffusion priors,'' \emph{arXiv preprint arXiv:2303.15181}, 2023.

\bibitem{zhang2023avatarverse}
H.~Zhang, B.~Chen, H.~Yang, L.~Qu, X.~Wang, L.~Chen, C.~Long, F.~Zhu, K.~Du, and M.~Zheng, ``Avatarverse: High-quality \& stable 3d avatar creation from text and pose,'' 2023.

\bibitem{Sketch2Scene2024}
\BIBentryALTinterwordspacing
XRVisionLabs, ``Sketch2scene: Automatic generation of interactive 3d game scenes from user's casual sketches,'' 2024. [Online]. Available: \url{https://xrvisionlabs.github.io/Sketch2Scene/}
\BIBentrySTDinterwordspacing

\bibitem{WorldSmith2023}
\BIBentryALTinterwordspacing
S.~Amershi, D.~S. Weld, M.~Vorvoreanu, A.~Fourney, B.~Nushi, P.~Collisson, J.~Suh, S.~T. Iqbal, P.~N. Bennett, K.~Inkpen, J.~T. Quinn, R.~Kikin-Gil, and E.~Horvitz, ``Worldsmith: Iterative and expressive prompting for world building with a generative ai,'' \emph{Proceedings of the 36th Annual ACM Symposium on User Interface Software and Technology}, 2023. [Online]. Available: \url{https://dl.acm.org/doi/10.1145/3586183.3606772}
\BIBentrySTDinterwordspacing

\bibitem{Magic3DSketch2024}
\BIBentryALTinterwordspacing
Y.~Zhang, Y.~Han, C.~Ding, J.~Zhang, and T.~Chen, ``Magic3dsketch: Create colorful 3d models from sketch-based 3d modeling guided by text and language-image pre-training,'' \emph{Neurocomputing}, 2024. [Online]. Available: \url{https://arxiv.org/pdf/2407.19225}
\BIBentrySTDinterwordspacing

\bibitem{zhu2025pcf}
R.~Zhu, S.~Qiu, Q.~Wu, K.-H. Hui, P.-A. Heng, and C.-W. Fu, ``Pcf-lift: Panoptic lifting by probabilistic contrastive fusion,'' in \emph{European Conference on Computer Vision}.\hskip 1em plus 0.5em minus 0.4em\relax Springer, 2025, pp. 92--108.

\bibitem{qiu2024advancing}
S.~Qiu, B.~Xie, Q.~Liu, and P.-A. Heng, ``Advancing extended reality with 3d gaussian splatting: Innovations and prospects,'' \emph{arXiv preprint arXiv:2412.06257}, 2024.

\bibitem{bdiot2023synthesizing}
Y.~Liang, X.~Yang, J.~Lin, H.~Li, X.~Xu, and Y.~Chen, ``Synthesizing 3d vr sketch using generative adversarial neural network,'' in \emph{BDIOT'23}, 2023.

\bibitem{arxiv2408vrcopilot}
L.~Zhang, J.~Pan, J.~Gettig, S.~Oney, and A.~Guo, ``Vrcopilot: Authoring 3d layouts with generative ai models in vr,'' \emph{arXiv preprint arXiv:2408.09382}, 2024.

\bibitem{vr2024text2vrscene}
Y.~Liang, X.~Yang, J.~Lin, H.~Li, X.~Xu, and Y.~Chen, ``Text2vrscene: Exploring the framework of automated text-driven generation system for vr experience,'' in \emph{IEEE Conference on Virtual Reality and 3D User Interfaces}, 2024.

\bibitem{arxiv2023speechtoimage}
Y.~Hu, M.~Yuan, K.~Xian, D.~S. Elvitigala, and A.~Quigley, ``Speech to image and text (ar): Exploring the design space of employing ai-generated content for augmented reality display,'' \emph{arXiv preprint arXiv:2303.16593}, 2023.

\bibitem{arxiv2020arsketch}
Y.~Liang, X.~Yang, J.~Lin, H.~Li, X.~Xu, and Y.~Chen, ``Arsketch: Sketch-based user interface for augmented reality,'' in \emph{UIST'20}, 2020.

\bibitem{vr2018fashiondesign}
------, ``A compensation method of two-stage image generation for human-ai collaborated in-situ fashion design in augmented reality environment,'' in \emph{IEEE Conference on Virtual Reality and 3D User Interfaces}, 2018.

\bibitem{tvcg2016liftoff}
B.~Jackson and D.~F. Keefe, ``Lift-off: Using reference imagery and freehand sketching to create 3d models in vr,'' in \emph{TVCG'16}, 2016.

\bibitem{unity2d3dpaint}
\BIBentryALTinterwordspacing
{Unity Technologies}, ``2d/3d paint,'' Accessed on 2023-MM-DD, 2023, unity Asset Store. [Online]. Available: \url{https://assetstore.unity.com/packages/tools/painting/2d-3d-paint-212475?srsltid=AfmBOopJrIk2WVuINSJFuwf6B3dg133OMuEqkh3MnjIaGL063hlVQbvy}
\BIBentrySTDinterwordspacing

\bibitem{mrtk}
MicroSoft, ``Mixed reality toolkit for unity,'' \url{https://github.com/MixedRealityToolkit/MixedRealityToolkit-Unity}, 2023.

\bibitem{meta_voice_sdk}
Meta, ``Voice sdk overview,'' \url{https://developers.meta.com/horizon/documentation/unity/voice-sdk-overview/}, 2023.

\bibitem{gatis2023rembg}
D.~Gatis, ``{rembg: Remove images background},'' \url{https://github.com/danielgatis/rembg}, 2023.

\bibitem{schaefer2005dual}
S.~Schaefer and J.~Warren, ``Dual marching cubes: Primal contouring of dual grids,'' Rice University, Houston, TX, Tech. Rep., 2005.

\bibitem{anonymous2023synthesizing}
W.~Li, ``Synthesizing 3d vr sketch using generative adversarial neural network,'' in \emph{Proceedings of the 2023 7th International Conference on Big Data and Internet of Things}, 2023, pp. 122--128.

\end{thebibliography}

\end{document}